# EVOLVING A STIGMERGIC
# SELF-ORGANIZED DATA-MINING

Vitorino Ramos
*CVRM-GeoSystems Centre,*
*IST, Technical University of Lisbon, PORTUGAL*
vitorino.ramos@alfa.ist.utl.pt

Ajith Abraham
*Natural Computation Lab, Department of Computer Science,*
*Oklahoma State University, Tulsa, OK, USA*
aa@cs.okstate.edu

**ABSTRACT**

Self-organizing complex systems typically are comprised of a large number of frequently similar components or events. Through their process, a pattern at the global-level of a system emerges solely from numerous interactions among the lower-level components of the system. Moreover, the rules specifying interactions among the system's components are executed using only local information, without reference to the global pattern, which, as in many real-world problems is not easily accessible or possible to be found. *Stigmergy*, a kind of indirect communication and learning by the environment found in social insects is a well know example of self-organization, providing not only vital clues in order to understand how the components can interact to produce a complex pattern, as can pinpoint simple biological non-linear rules and methods to achieve improved artificial intelligent adaptive categorization systems, critical for Data-Mining. On the present work it is our intention to show that a new type of Data-Mining can be designed based on *Stigmergic* paradigms, taking profit of several natural features of this phenomenon. By hybridizing bio-inspired Swarm Intelligence with Evolutionary Computation we seek for an entire distributed, adaptive, collective and cooperative self-organized Data-Mining. As a real-world / real-time test bed of our proposal, World-Wide-Web Mining will be used. Having that purpose in mind, Web usage Data was collected from the *Monash University's* Web site (Australia), with over 7 million hits every week. Results are compared to other recent systems, showing that the system presented is by far promising.

## 1. INTRODUCTION

Self-Organization refers to a broad range of pattern-formation processes in both physical and biological systems, such as sand grains assembling into rippled dunes, chemical reactants forming swirling spirals, cells making up highly structured tissues, and fish joining together in schools. A basic feature of these diverse systems is the means by which they acquire their order and structure. In self-organizing systems, pattern formation occurs through interactions internal to the system, without intervention by external directing influences. As defined by *Camazine* et al [29], self-organization is a process in which pattern at the global level of a system emerges solely from numerous interactions among the lower-level components of the system. Moreover, the rules specifying interactions among the system's components are executed using only local information, without reference to the global pattern.

One well know example is provided by the emergence of self-organization in social insects, via direct (mandibular, antennation, chemical or visual contact, etc) or indirect interactions. The latter types are more subtle and defined by *Grassé* as stigmergy [11] to explain task coordination and regulation in the context of nest reconstruction in *Macrotermes* termites. An example [10], could be provided by two individuals, who interact indirectly when one of them modifies the environment and the other responds to the new environment at a later time. In other words, stigmergy could be defined as a typical case of environmental synergy or learning via the environment. *Grassé* showed that the coordination and regulation of building activities do not depend on the workers themselves but are mainly achieved by the nest structure: a stimulating configuration triggers the response of a termite worker, transforming the configuration into another configuration that may trigger in turn another (possibly different) action performed by the same termite or any other worker in the colony. Another illustration of how stimergy and self-organization can be combined into more subtle adaptive behaviors is recruitment in social insects. Self-organized trail laying by individual ants is a way of modifying the environment to communicate with nest mates that follow such trails. It appears that task performance by some workers decreases the need for more task performance: for instance, nest cleaning by some workers reduces the need for nest cleaning [10,9]. Therefore, nest mates communicate to other nest mates by modifying the environment (cleaning the nest), and nest mates respond to the modified environment (by not engaging in nest cleaning); that is stigmergy. Division of labor is another paradigmatic phenomenon of stigmergy. But by far more crucial to the present work and aim, is how ants form piles of items such as dead bodies (corpses), larvae, or grains of sand (fig.1a, section 3.1). There again, stigmergy is at work: ants deposit items at initially random locations. When other ants perceive deposited items, they are stimulated to deposit items next to them, being this type of cemetery clustering organization and brood sorting a type of self-organization and adaptive behavior.

There are other types of examples (e.g. prey collectively transport), yet stigmergy is also present: ants change the perceived environment of other ants (their cognitive map, according to [6,28]), and in every example, the environment serves as medium of communication. What all these examples have in common is that they show how stigmergy can easily be made operational. As mentioned by *Bonabeau* et al. [10], that is a promising first step to design groups of artificial agents which solve problems: replacing coordination (and possible some hierarchy) through direct communications by indirect interactions is appealing if one wishes to design simple agents and reduce communication among agents. Finally, stigmergy is often associated with flexibility: when the environment changes because of an external perturbation, the insects respond *appropriately* to that perturbation, as if it were a modification of the environment caused by the colony's activities. In other words, the colony can collectively respond to the perturbation with individuals exhibiting the same behavior. When it comes to artificial agents, this type of flexibility is priceless: it means that the agents can respond to a perturbation without being reprogrammed to deal with that particular instability. In our context, this means that no classifier re-training is needed for any new sets of data-item types (new classes) arriving to the system, as is necessary in many classical models, or even in some recent ones. Moreover, the data-items that were used for supervised purposes in early stages in the colony evolution in his exploration of the search-space, can now, along with new items, be re-arranged in more optimal ways. Classification and/or data retrieval remains the same, but the system organizes itself in order to deal with new classes, or even new sub-classes. This task can be performed in real time, and in robust ways due to system's redundancy, as was shown in [23,2].

On the present work it is our intention to show that a new type of Data-Mining can be designed based on Stigmergic paradigms, taking profit of the above cited different natural features, incorporating them on the intelligent processing. By hybridizing Swarm Intelligence with Evolutionary Computation we seek for an entire distributed, adaptive, collective and cooperative self-organized Data-Mining. As a real-world / real-time test bed of our proposal, World-Wide-Web Mining will be used.

## 2. STIMERGY, SELF-ORGANIZATION AND DATA-MINING

Data clustering is also one of those problems in which real ants can suggest very interesting heuristics for computer scientists. One of the first studies using the metaphor of ant colonies related to the above clustering domain is due to *Deneubourg* [8], where a population of ant-like agents randomly moving onto a 2D grid are allowed to move basic objects so as to cluster them. This method was then further generalized by *Lumer* and *Faieta* [18] (here after *LF* algorithm), applying it to exploratory data analysis, for the first time. In 1995, the two authors were then beyond the simple example, and applied their algorithm to interactive exploratory database analysis, where a human observer can probe the contents of each represented point (sample, image, item) and alter the characteristics of the clusters. They showed that their model provides a way of exploring complex information spaces, such as document or relational databases, because it allows information access based on exploration from various perspectives. However, this last work entitled "Exploratory Database

Analysis via Self-Organization", according to [10], was never published due to commercial applications. They applied the algorithm to a database containing the "profiles" of 1650 bank customers. Attributes of the profiles included marital status, gender, residential status, age, a list of banking services used by the customer, etc. Given the variety of attributes, some of them qualitative and others quantitative, they had to define several dissimilarity measures for the different classes of attributes, and to combine them into a global dissimilarity measure (in, pp. 163, Chapter 4 [10]). More recently, *Ramos* et al. [26,25,28] presented a novel strategy (ACLUSTER) to tackle unsupervised clustering as well as data retrieval problems, avoiding not only short-term memory based strategies, as well as the use of several artificial ant types (using different speeds), present in those approaches proposed initially by *Lumer* [18]. Other works in this area include those from *Monmarché* et al. (1999 - [21]), *Ramos*, *Merelo* et al. (2000-02 – [28,25,26,27]), *Hoe* et al. (2002 - [14]), *Handl* and *Dorigo* (2003 – [12]) *Ramos* and *Abraham* (2003-04 – [23,2,1]). Applications range from Data Clustering and Exploratory Data Analysis (all the above references) to Document Processing and Analysis [26,14,12], Image Retrieval systems [25], Continuous Mappings and Classification [23], Perception and Memory on Digital Images [28], Collective Robotics [8,24], Architecture and Art [27], as well as to World-Wide-Web usage Mining [2] and Network Management [1]. Finally, and even if not related to clustering, other researchers as *Parpinelli* et al [22] approached Data-Mining via ACO, the famous *Dorigo's* Ant Colony Optimization Algorithm [10,9], extracting classification rules from data, having in mind an optimized framework.

## 3. WORLD-WIDE-WEB MINING

Web mining is the extraction of interesting and useful knowledge and implicit information from artifacts or activity related to the WWW [17,7]. Web servers record and accumulate data about user interactions whenever requests for resources are received. Analyzing the Web access logs can help understand the user behavior and the web structure. From the business and applications point of view, knowledge obtained from the Web usage patterns could be directly applied to efficiently manage activities related to e-business, e-services, e-education, on-line communities and so on. Accurate Web usage information could help to attract new customers, retain current customers, improve cross marketing/sales, effectiveness of promotional campaigns, track leaving customers and find the most effective logical structure for their Web space. User profiles could be built by combining users' navigation paths with other data features, such as page viewing time, hyperlink structure, and page content [13]. However as the size and complexity of the data increases, the statistics provided by existing Web log file analysis tools may prove inadequate and more intelligent mining techniques will be necessary. *Jespersen* et al [15] proposed a hybrid approach for analyzing the visitor click sequences. A combination of hypertext probabilistic grammar and click fact table approach is used to mine Web logs which could be also used for general sequence mining tasks. *Mobaster* et al [20] proposed the Web personalization system which consists of offline tasks related to the mining of usage data and online process of automatic Web page customization based on the knowledge discovered. LOGSOM proposed by *Smith* et al [30], utilizes a self-organizing map (SOM) to organize web pages into a two-dimensional map based solely on the users' navigation behavior, rather than the content of the web pages. "LumberJack" proposed by *Chi* et al [5] builds up user profiles by combining both user session clustering and traditional statistical traffic analysis using K-means algorithm. Also using SOM, *Merelo* et al [19] proposed the clustering and mapping of Web-log communities. *Joshi* et al [16] used relational online analytical processing approach for creating a Web log warehouse using access logs and mined logs (association rules and clusters). A comprehensive overview of Web usage mining research is found in [7,31].

In this paper, an ant colony clustering (ACLUSTER) [25] is proposed to segregate visitors and thereafter a linear genetic programming approach [4] to analyze the visitor trends. The results are compared with the earlier works using self organizing maps [32] and evolutionary-fuzzy C-means algorithm [3] to segregate the user access records and several soft computing paradigms to analyze the user access trends. Web access log data at the *Monash* University's Web site [34] were used for experimentations. The University's central web server receives over 7 million hits in a week and therefore it is a real challenge to find and extract hidden usage pattern information. Average daily and hourly access patterns for 5 weeks (11 August'02 – 14 September'02) were used. The average daily and hourly patterns nevertheless tend to follow a similar trend (as evident from the figures) the differences tend to increase during high traffic days (Monday – Friday) and during the peak hours (11:00-17:00 Hrs). Due to the enormous traffic volume and chaotic access behavior, the prediction of the user access patterns becomes more difficult and complex. On the present work we propose a self-organized and adaptive-evolving approach for this specific data-mining problem. In the subsequent section, the proposed architecture is presented as experimentation results of the ant clustering – linear genetic programming approach. Some conclusions are provided towards the end.

## 4. THE ANT COLONY BASED CLUSTERING AND LINEAR GENETIC PROGRAMMING APPROACH (ANT-LGP)

The hybrid framework uses an ant colony optimization algorithm to cluster Web usage patterns. The raw data from the log files are cleaned and pre-processed and the ACLUSTER algorithm [25] is used to identify the usage patterns (data clusters). Each data-item is considered as an object, which will be transported by the artificial ants on a 2D classification space. The developed clusters of data are fed to a linear genetic programming model to analyze the usage trends.

### 4.1 Distributed, collaborative and Stigmergic clustering

The swarm intelligence algorithm fully uses agents that stochastically move around the classification "habitat" following pheromone concentrations. That is, instead of trying to solve some disparities in the basic LF algorithm by adding different ant casts, short-term memories and behavioral switches, which are computationally intensive, representing simultaneously a potential and difficult complex parameter tuning, it was our intention to follow real ant-like behaviors as possible (some other features will be incorporated, as the use of different response thresholds to task-associated stimulus intensities, discussed later). In that sense, bio-inspired spatial transition probabilities are incorporated into the system, avoiding randomly moving agents, which tend the distributed algorithm to explore regions manifestly without interest (e.g., regions without any type of object clusters), being generally, this type of exploration, counterproductive and time consuming. Since this type of transition probabilities depend on the spatial distribution of pheromone across the environment, the behavior reproduced is also a stigmergic one. Moreover, the strategy not only allows to guide ants to find clusters of objects in an adaptive way (if, by any reason, one cluster disappears, pheromone tends to evaporate on that location), as the use of embodied short-term memories is avoided (since this transition probabilities tends also to increase pheromone in specific locations, where more objects are present). As we shall see, the distribution of the pheromone represents the memory of the recent history of the swarm, and in a sense it contains information which the individual ants are unable to hold or transmit. There is no direct communication between the organisms but a type of indirect communication through the pheromonal field. In fact, ants are not allowed to have any memory and the individual's spatial knowledge is restricted to local information about the whole colony pheromone density. In order to design this behavior, one simple model was adopted (*Chialvo* and *Millonas*, [6]), and extended (as in [28]) due to specific constraints of the present proposal. As described in [6], the state of an individual ant can be expressed by its position *r*, and orientation *q*. It is then sufficient to specify a transition probability from one place and orientation $(r,q)$ to the next $(r^*,q^*)$ an instant later. The response function can effectively be translated into a two-parameter transition rule between the cells by use of a pheromone weigthing function (Eq. 4.1). This equation measures the relative probabilities of moving to a cite *r* (in our context, to a grid location) with pheromone density *s*(*r*). The parameter *b* is associated with the osmotropotaxic sensitivity (a kind of instantaneous pheromonal gradient following), and on the other hand, 1/*d* is the sensory capacity, which describes the fact that each ant's ability to sense pheromone decreases somewhat at high concentrations. In addition to the former equation, there is a weigthing factor $w(Dq)$, where *Dq* is the change in direction at each time step, i.e. measures the magnitude of the difference in orientation. As an additional condition, each individual leaves a constant amount *h* of pheromone at the cell in which it is located at every time step *t*.

$$W(s) = \left(1 + \frac{s}{1+ds}\right)^b \quad (4.1) \qquad P_{ik} = \frac{W(s_i)w(\Delta_i)}{\sum_{j/k} W(s_j)w(\Delta_j)} \quad (4.2)$$

**Algorithm.** High-level description of *ACLUSTER*.

/* **Initialization** */
**For** every object or data-item $o_i$ **do**
Place $o_i$ randomly on grid
**End For**
**For** all agents **do**
Place agent at randomly selected site
**End For**
/* **Main loop** */
**For** $t = 1$ to $t_{max}$ **do**
**For** all agents **do**
$sum = 0$
Count the number of items $n$ around site $r$
**If** ((agent unladen) and (site $r$ occupied by item $o_i$)) **then**
**For** all sites around $r$ with items present **do**
/* **According to Eqs. 4.4, 4.6 and Table 1 (4.1.1)** */
Compute $d, \chi, e$ and $P_p$
Draw a random real number $R$ between 0 and 1
**If** ($R = P_p$) **then**
$sum = sum + 1$
**End If**
**End For**
**If** (($sum = n/2$) or ($n = 0$)) **then**
Pick up item $o_i$
**End If**
**Else If** ((agent carrying item $o_i$) and (site $r$ empty)) **then**
**For** all sites around $r$ with items present **do**
/* **According to Eqs. 4.4, 4.5 and Table 1 (4.1.1)** */
Compute $d, \chi, d$ and $P_d$
Draw random real number $R$ between 0 and 1
**If** ($R = P_d$) **then**
$sum = sum + 1$
**End If**
**End For**
**If** ($sum = n/2$) **then**
Drop item $o_i$
**End If**
**End If**
/* **According to Eqs. 4.1 and 4.2 (section 4.1)** */
**Compute** $W(s)$ and $P_{ik}$
**Move** to a selected $r$ not occupied by other agent
**Count** the number of items $n$ around that new site $r$
**Increase** pheromone at site $r$ according to $n$, that is:
$P_r = P_r + [h + (n/a)]$
**End For**
**Evaporate** pheromone by $K$, at all grid sites
**End For**
**Print** location of items

/* **Values of parameters used in experiments** */
$k_1 = 0.1$, $k_2 = 0.3$, $K = 0.015$, $h = 0.07$, $a = 400$,
$b = 3.5$, ?$=0.2$, $t_{max} = 10^6$ steps.

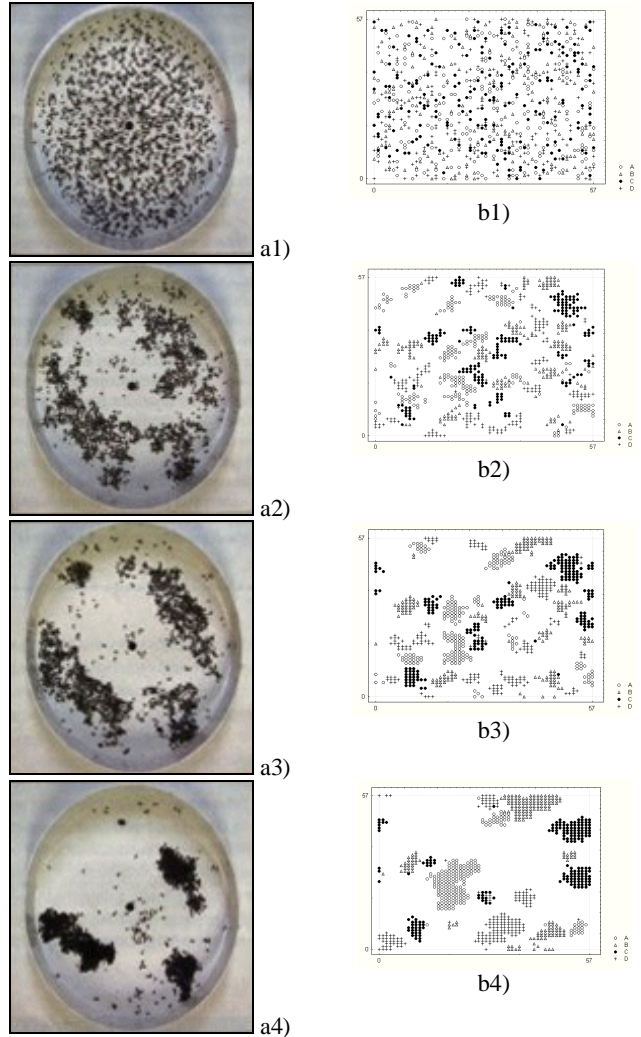

a1) a2) a3) a4)

b1) b2) b3) b4)

Fig. 1 – From a1) to a4), a sequential clustering task of corpses performed by a real ant colony. 1500 corpses are randomly located in a circular arena with radius = 25 cm, where *Messor Sancta* ant workers are present. The figure shows the initial state (a1), 2 hours (a2), 6 hours (a3) and 26 hours (a4) after the beginning of the experiment [10]. In b1-b4, some experiments with the present algorithm, conducted on synthetic data (as in [25,18]). Spatial distribution of 800 items on a 57 x 57 non-parametric toroidal grid at several time steps. At $t=1$, four types of items are randomly allocated into the grid. As time evolves, several homogenous clusters emerge due to the ant colony action, and as expected the global entropy decreases [26].

This pheromone decays at each time step at a rate $k$. Then, the normalised transition probabilities on the lattice to go from cell $k$ to cell $i$ are given by $P_{ik}$ [6] (Eq. 4.2), where the notation $j/k$ indicates the sum over all the pixels $j$ which are in the local neighbourhood of $k$. $D_i$ measures the magnitude of the difference in orientation for the previous direction at time $t-1$.

### 4.1.1 Picking and dropping data-objects

In order to model the behavior of ants associated to different tasks, as dropping and picking up objects, we suggested [28] the use of combinations of different response thresholds. As we have seen before, there are two major factors that should influence any local action taken by the ant-like agent: the number of objects in his neighborhood, and their similarity (including the hypothetical object carried by one ant). *Lumer* and *Faieta* [18], use an average similarity, mixing distances between objects with their number, incorporating it simultaneously into a response threshold function. Instead, we recommend the use of combinations of two independent response threshold functions, each associated with a different environmental factor (or, stimuli intensity), that is, the number of objects in the area, and their similarity. Moreover, the computation of average similarities are avoided in the present algorithm, since this strategy can be somehow blind to the number of objects present in one specific neighborhood. In fact, in *Lumer* and *Faieta*'s work [18], there is an hypothetical chance of having the same average similarity value, respectively having one or, more objects present in that region. But, experimental evidences and observation in some types of ant colonies, can provide us with a different answer. After *Wilson* (*The Insect Societies*, Cambridge Press, 1971), it is known that minors and majors in the polymorphic species of ants *Genus Pheidole*, have different response thresholds to task-associated stimulus intensities (i.e., division of labor). Recently, and inspired by this experimental evidence, *Bonabeau* et al. [10], proposed a family of response threshold functions in order to model this behavior. According to it, every individual has a response threshold $q$ for every task. Individuals engage in task performance when the level of the task-associated stimuli $s$, exceeds their thresholds. Author's defined $s$ as the intensity of a stimulus associated with a particular task, i.e. $s$ can be a number of encounters, a chemical concentration, or any quantitative cue sensed by individuals. One family of response functions $T_q(s)$ (the probability of performing the task as a function of stimulus intensity $s$), that satisfy this requirement is given by (Eq. 4.3) [10], where $n>1$ determines the steepness of the threshold (normally $n=2$, but similar results can be obtained with other values of $n>1$). Now, at $s = q$, this probability is exactly ½. Therefore, individuals with a lower value of $q$ are likely to respond to a lower level of stimulus. In order to take account on the number of objects present in one neighborhood, Eq. 4.3, was used (where, $n$ now stands for the number of objects present in one neighborhood, and $q = 5$), defining $\chi$ (Eq. 4.4) as the response threshold associated to the number of items present in a 3 x 3 region around $r$ (one specific grid location). Now, in order to take account on the hypothetical similarity between objects, and in each ant action due to this factor, a Euclidean normalized distance $d$ is computed within all the pairs of objects present in that 3 x 3 region around $r$. Being $a$ and $b$, a pair of objects, and $f_a(i)$, $f_b(i)$ their respective feature vectors (being each object defined by $F$ features), then $d = (1/d_{max}).[(1/F).\sum_{i=1,F}(f_a(i)-f_b(i))^2]^{½}$. Clearly, this distance $d$ reaches its maximum (=1, since $d$ is normalized by $d_{max}$) when two objects are maximally different, and $d=0$ when they are equally defined by the same $F$ features. Moreover, $d$ and $e$ (Eqs. 4.5, 4.6), are respectively defined as the response threshold functions associated to the similarity of objects, in case of dropping an object (Eq. 4.5), and picking it up (Eq. 4.6), at site $r$. Finally, in every action taken by an agent, and in order to deal, and represent different stimulus intensities (number of items and their similarity), present at each site in the environment visited by one ant, the strategy used a composition of the above defined response threshold functions (Eqs. 4.4, 4.5 and 4.6). Several composed probabilities were analyzed [25] and used as test functions in one preliminary test. The best results were achieved with the test function #1 below (table 1), achieving a high classification rate (out of 4 different functions were used, as well the *LF* algorithm [18] for comparison reasons – see [26,25]). Alternatively, the system can also be robust feeding the data continuously (for instance, as they appear) as proved in past works [23]. For other algorithm details please consult [25,26,28].

$$T_q(s) = \frac{s^n}{s^n + q^n} \quad (4.3) \qquad c = \frac{n^2}{n^2 + 5^2} \quad (4.4)$$

$$d = \left(\frac{k_1}{k_1 + d}\right)^2 \quad (4.5) \qquad e = \left(\frac{d}{k_2 + d}\right)^2 \quad (4.6)$$

Table 1. Type #1 Probability functions used (as in [25,26])

| Picking Probability | Dropping Probability |
|---|---|
| $P_p = (1-c).e$ | $P_d = c.d$ |

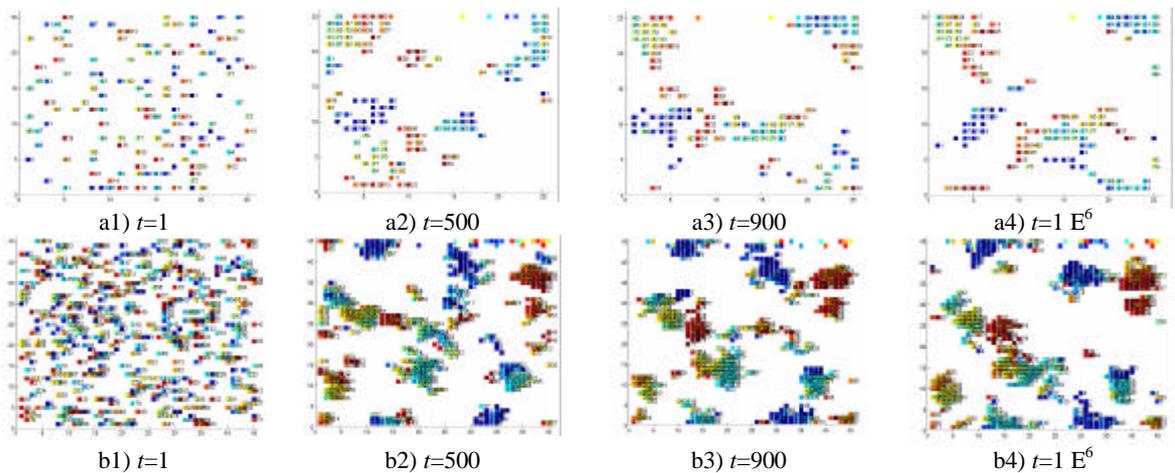

Fig. 2 – From a1) to a4) the snapshots represent the spatial distribution of daily Web traffic data on a 25 x 25 non-parametric toroidal grid at several time steps. At $t=1$, data items are randomly allocated into the grid. As time evolves, several homogenous clusters emerge due to the ant colony action. Type #1 [26] probability function was used with $k_1=$ 0.1, $k_2=$ 0.3 and 14 ants. ($k_1$ and $k_2$ are threshold constants). From b1) to b4), similar results for the hourly Web traffic data, on a 45 x 45 non-parametric toroidal grid with 48 ants present (further cluster mappings can be found on [2]).

## 4.2 Experimental setup and Clustering results

In this research, we used the statistical/ text data generated by the log file analyzer from January 1st, 2002 to July 7th, 2002. Selecting useful data is an important task in the data pre-processing block. After some preliminary analysis, we selected the statistical data comprising of domain byte requests, hourly page requests and daily page requests as focus of the cluster models for finding Web users' usage patterns. The most recently accessed data were indexed higher while the least recently accessed data were placed at the bottom. For each of datasets (daily and hourly log data), the algorithm was run twice (for $t=1$ to $1E^6$) in order to check if somehow the results were similar (which they appear to be, if we look into which data items are connected into what clusters). The classification space is always 2D, non-parametric and toroidal (see section 4.1). Experimentation results for the daily and hourly Web traffic data are accessible in figure 2-a) and 2-b).

## 4.3 The Linear Genetic Programming (LGP) approach

Linear genetic programming is a variant of the GP technique that acts on linear genomes [4]. Its main characteristics in comparison to tree-based GP lies in that the evolvable units are not the expressions of a functional programming language (like LISP), but the programs of an imperative language (like c/c ++). An alternate approach is to evolve a computer program at the machine code level, using lower level representations for the individuals. This can tremendously hasten the evolution process as, no matter how an individual is initially represented, finally it always has to be represented as a piece of machine code, as fitness evaluation requires physical execution of the individuals.

The basic unit of evolution here is a native machine code instruction that runs on the floating-point processor unit (FPU). Since different instructions may have different sizes, here instructions are clubbed up together to form instruction blocks of 32 bits each. The instruction blocks hold one or more native machine code instructions, depending on the sizes of the instructions. A crossover point can occur only between instructions and is prohibited from occurring within an instruction. However the mutation operation does not have any such restriction.

## 4.4 Experimental setup and Trend Analysis results

Besides the inputs *'volume of requests'* and *'volume of pages (bytes)'* and *'index number'*, we also used the *'cluster information'* provided by the clustering algorithm as an additional input variable. The data was re-indexed based on the cluster information. Our task is to predict (a few time steps ahead) the Web traffic volume on a hourly and daily basis. We used the data from 17 February 2002 to 30 June 2002 for training and the data from 01 July 2002 to 06 July 2002 for testing and validation purposes. We used a LGP technique that manipulates and evolves a program at the machine code level. We used the *Discipulus* workbench for simulating LGP [33]. The settings of various linear genetic programming system parameters are of utmost importance for successful performance of the system. The population space has been subdivided into multiple subpopulation or demes. Migration of individuals among the subpopulations causes evolution of the entire population. It helps to maintain diversity in the population, as migration is restricted among the demes. Moreover, the tendency towards a bad local minimum in one deme can be countered by other demes with better search directions. The various LGP search parameters are the mutation frequency, crossover frequency and the reproduction frequency: The crossover operator acts by exchanging sequences of instructions between two tournament winners. Steady state genetic programming approach was used to manage the memory more effectively. After a trial and error approach, the parameter settings used for the experiments were: *Population size:* 500, *Tournament size:* 4, *Maximum no. of tournaments:* 120.000, *Mutation frequency:* 90%, *Crossover frequency:* 80%, *Number of demes:* 10, *Maximum program size:* 512 and *Target subset size:* 100.

These experiments were repeated three times and the test data was passed through the saved model. Figures 3 illustrates the average growth in program length for hourly Web traffic, as well as depict the training and test performance (average training and test fitness) for hourly Web traffic. Similar results on the average growth as well as for the test performance were found for the daily Web traffic data [2]. Empirical comparisons of the proposed framework with some of our previous works are depicted in Tables 2 and 3. These tables show a comparison of the proposed framework (ANT-LGP) with *i-Miner* (hybrid evolutionary fuzzy clustering–fuzzy inference system) [3], self-organizing map–linear genetic programming (SOM-LGP) [32] and self-organizing map–artificial neural network (SOM-ANN) [32].

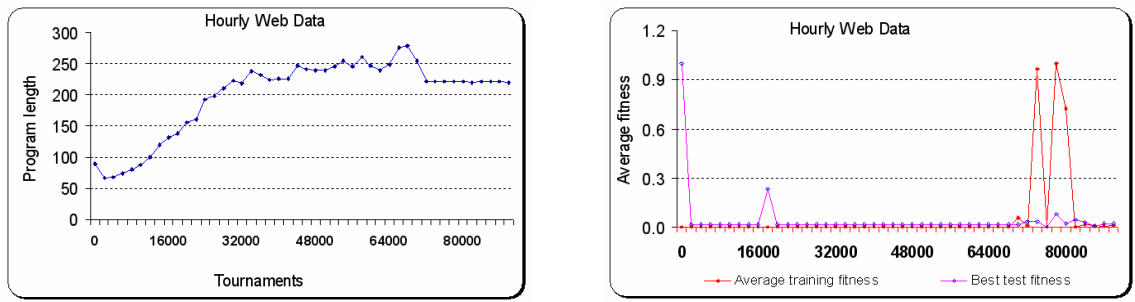

Fig 3. – Hourly Web data analysis: On the left growth in average program length during 120000 tournaments, and on the right comparison of average training and test fitness during 120,000 tournaments (supplementary results in [2]).

Table 2. Performance of the different paradigms for the daily Web data period

| Hybrid method | Daily (1 day ahead) | | |
|---|---|---|---|
| | RMSE | | CC |
| | Train | Test | |
| **ANT-LGP** | 0.0191 | 0.0291 | 0.9963 |
| ***i-Miner* (FCM-FIS)** | 0.0044 | 0.0053 | 0.9967 |
| **SOM-ANN** | 0.0345 | 0.0481 | 0.9292 |
| **SOM-LGP** | 0.0543 | 0.0749 | 0.9315 |

Table 3. Performance of the different paradigms for the hourly Web data period

| Hybrid method | Hourly (1 hour ahead) | | |
|---|---|---|---|
| | RMSE | | CC |
| | Train | Test | |
| **ANT-LGP** | 0.2561 | 0.035 | 0.9921 |
| ***i-Miner* (FCM-FIS)** | 0.0012 | 0.0041 | 0.9981 |
| **SOM-ANN** | 0.0546 | 0.0639 | 0.9493 |
| **SOM-LGP** | 0.0654 | 0.0516 | 0.9446 |

## 5. CONCLUSION

We have presented in this paper a new evolutionary and self-organized bio-inspired shift for Data-Mining, unsupervised clustering and data exploratory analysis, while sketching a clear parallel between a mode of problem-solving in social insects and a distributed, reactive, algorithmic approach. Some of the mechanisms underlying corpse clustering, brood sorting and those that can explain the worker's behavioral flexibility, as regulation of labor and allocation of tasks have first been introduced. As in similar past works applied to document clustering and text retrieval, the role of response thresholds to task-associated stimulus intensities were stressed as an important part of the strategy, and incorporated into the algorithm by using compositions of different response functions. These compositions allows the strategy not only to be more accurate relatively to behaviors found in nature as avoids short-term memory based strategies, and the use of several artificial ant types (using different speeds), present in some recent approaches. Behavioral switches as used in [18], were also avoided, in order to maintain simplicity and to avoid complex parameter settings to be performed by the domain expert. At the level of agent moves in the grid, a truly stigmergic model was introduced (section 4) in order to deal with clusters of objects, avoiding randomly moves which can be counterproductive in the distributed search performed by the swarm, and adopted by all past models. In fact, the present algorithm, along with [26,25], were the first to introduce pheromone traces on agents to deter random explorations and encourage objects cluster formation, a successful feature not implemented even in some recent proposals [14]. In here however, while achieving similar results compared to the *LF* model [18], as pointed by the spatial entropy of solutions at each iteration, the present algorithm is by far simpler. Moreover, for some of response thresholds compositions used, results are superior while using the present algorithm for the majority of time iterations [25], that is, entropy is always lower, even if at the end they tend to the same value. As a final advantage, the present framework does not require any initial information about the future classification, such as an initial partition or an initial number of classes.

This novel evolutionary strategy was then applied for the first time to Web Mining by means of the hybridization with Linear Genetic programming (LGP). The proposed ANT-LGP model seems to work very well for the problem considered. The empirical results also reveal the importance of using optimization techniques for mining useful information. In this paper, our focus was to develop accurate trend prediction models to analyze the hourly and daily web traffic volume. Useful information could be discovered from the clustered data. The knowledge discovered from the developed clusters using different intelligent models could be a good comparison study and is left as a future research topic. As illustrated in Tables 2 and 3, incorporation of the ant clustering algorithm helped to improve the performance of the LGP model (when compared to clustering using self organizing maps). *i-Miner* framework gave the overall best results with the lowest RMSE on test error and the highest correlation coefficient (CC).

Finally, and as verified by other tests [23] on parts of the present framework (ACLUSTER), a robust nonstop classifier could be achieved, which produces class decisions on a continuous stream data, allowing for continuous mappings. As we know, many categorization systems have the inability to perform classification and visualization in a continuous basis or to self-organize new data-items into the older ones (even more, into new labels if necessary), unless a new training happens. This disadvantage is also present in more recent approaches using Self-Organizing Maps, as in *Kohonen* maps. While a benchmark comparison of the above cited methods should be interesting and necessary to explore, the ability of ACLUSTER to perform continuous mappings and the incapacity of the latter to conceive it, tend to difficult any serious comparison. Besides this limitation the present model expands our view of the problems we face. Future research will also incorporate more data mining algorithms to improve knowledge discovery and association rules from the clustered data. The contribution of the individual input variables and different clustering algorithms will be also investigated to improve the trend analysis and knowledge discovery. In these ways we believe it will help us find new kinds of World-Wide-Web community patterns we never thought of before.